\documentclass{article}
\usepackage{PRIMEarxiv}
\usepackage[utf8]{inputenc} 
\usepackage[T1]{fontenc}    
\usepackage{hyperref}       
\usepackage{url}            
\usepackage{booktabs}       
\usepackage{amsfonts}       
\usepackage{nicefrac}       
\usepackage{microtype}      
\usepackage{lipsum}
\usepackage{listings}

\usepackage{fancyhdr}       
\usepackage{graphicx}       
\graphicspath{{media/}}     

\usepackage{subcaption}
\usepackage{calc}
\usepackage{amssymb}
\usepackage{amstext}
\usepackage{amsmath}
\usepackage{amsthm}
\usepackage{multicol}
\usepackage{pslatex}
\usepackage{apalike}
\usepackage{algorithm2e}
\usepackage[bottom]{footmisc}

\usepackage{multirow}
\usepackage{graphicx}
\usepackage{xcolor}
\usepackage{verbatim}

\title{A First Context-Free Grammar Applied to Nawatl\\ Corpora Augmentation
\thanks{\textit{\underline{Citation}}: 
\textbf{Guzman-Landa et al. A First Context-Free Grammar Applied to Nawatl Corpora Augmentation. 11 pages.}} 
}

\author{Juan-José Guzm\'an-Landa, Juan-Manuel Torres-Moreno\\
Laboratoire Informatique d'Avignon, Avignon Université\\ BP 84911 Agroparc Cedex 9, Avignon, France\\
\texttt{\{juan-jose.guzman-landa, juan-manuel.torres\}@univ-avignon.fr}
\AND
Miguel Figueroa-Saavedra$^1$, Ligia Quintana-Torres$^2$, Martha-Lorena Avenda\~no-Garrido$^2$\\ 
$^2$Facultad de Matem\'aticas, $^1$Instituto de Investigaciones en Educación,\\
Universidad Veracruzana, Xalapa, Mexico\\
\texttt{\{migfigueroa, ligiaquintana, maravendano\}@uv.mx}
\AND
Graham Ranger\\
Laboratoire Identités Culturelles, Textes et Théâtralité, Avignon Université,\\
74 Rue Louis Pasteur, 84029 Avignon, France
\\ \texttt{graham.ranger@univ-avignon.fr}
}

\begin{document}
\maketitle

\begin{abstract}
In this article we introduce a context-free grammar (CFG) for the Nawatl language. Nawatl  is an Amerindian language of the $\pi$-language type, i.e. a language with few digital resources, in which the corpora available for machine learning are virtually non-existent. The objective here is to generate a significant number of grammatically correct artificial sentences, in order to increase the corpora available for language model training. We want to show that a grammar enables us significantly to expand a corpus in Nawatl which we call $\pi$-\textsc{yalli}.
The corpus, thus enriched, enables us to train algorithms such as FastText and to evaluate them on sentence-level semantic tasks. Preliminary results show that by using the grammar, comparative improvements are achieved over some LLMs. However, it is observed that to achieve more significant improvement, grammars that model the Nawatl language even more effectively are required.
\end{abstract}



\section{Introduction}
\label{sec:introduction}

Nawatl (or Nahuatl) is one of the indigenous languages of Mexico and the most widely spoken national indigenous language, with approximately 1,65 M speakers \cite{inegi2020censo}. The language is marked by substantial dialectal diversity, comprising 29 recognized variants distributed across four main regions: the Western, Central, Eastern, and Huastec zones (Ethnologue, 2025\footnote{\url{https://www.ethnologue.com}}; \cite{Lastra1986areas}).
This linguistic diversity presents challenges for the use of textual corpora in educational, communicative, and digital contexts, as it entails significant variation in orthography and lexical choice \cite{zimmerman},\cite{olko2016bridging},\cite{hansen2024nahuatl}, \cite{Figueroa2024amapowalistli}.

Although the publication of digital material in Nawatl has increased, the notable dispersion and variety of such resources has prevented them from gaining visibility on social media and from being clearly identified and accessed in repositories. This is nonetheless not an obstacle to the gradual increase in their presence and literate use, even if the digital linguistic resources available, which are essential for the current revitalization of the language, remain limited.\cite{pugh-etal-2025-ihquin}.

A serious problem is, therefore, the scarcity of computational resources for this language and, in particular, corpora available for machine learning. Our approach to mitigating this problem proposes the generation of artificial corpora that respect the structure of Nawatl using formal grammars.

Our goal in proposing the grammar is not to model the complexity of the Nawatl language fully, but rather to generate syntactically valid sentences. These sentences can serve as a basis for the creation of large-scale synthetic corpora.
Such corpora may be leveraged to enhance the training of both static and dynamic Large Language Models (LLMs) for $\pi$-languages —i.e. languages with limited digital resources \cite{these-pi,abdillahi:hal-01311495}. 
Specifically, we aim to expand the Nawatl corpus $\pi$-\textsc{yalli}\footnote{\url{https://demo-lia.univ-avignon.fr/pi-yalli/}} \cite{torres2024pi,NAHU2,piyalliTALN}, which has previously been employed in word-level semantic similarity tasks. 
 The size of the vocabulary has a direct impact during training: a larger vocabulary is expected to positively affect the performance of the model \cite{tunstall2022transformers,goyal2018deep}.

The structure of the paper is as follows: Section \ref{sec:nawatl} provides an overview of the Nawatl language and its grammatical features. Section \ref{sec:GNC} introduces context-free grammars. Section \ref{sec:GNC_proposed} presents the proposed context-free micro-grammar for Nawatl. Section \ref{sec:artificial_corpus} describes the extended corpus, $\pi$-yall-\textsc{ia}. Section \ref{sec:experiments} reports on experiments using the micro-grammars in a semantic similarity task. Finally, Section \ref{sec:conclusions_future} concludes the paper and outlines directions for future work.

\section{Nawatl language}
\label{sec:nawatl}

Nawatl, an Uto-Aztecan language, is an agglutinative and polysynthetic language that composes words by joining various morphemes to a verbal or nominal root, thereby constructing meaning.

Nawatl sentences have a basic verb–subject–object (VSO) syntax, but allows for structural flexibility that responds to speakers' needs. Thus, orders such as VO, VS, VOS, and, less frequently, SV, SVO, and SOV can also be found (Table \ref{tab:EXEMP_GRAM}). Syntactic relationships between words and clauses are established through the valency of the verb and the use of connectors (particles).

These connectors can also be formed through groupings that establish nuances of meaning, in addition to discourse connectors. Some of these words can form ``one-word sentences'', since their morphology includes the subject and predicate, as well as information about the actants and modal, directional, and relational elements (see \cite{Launey1978introduction}, \cite{floresnajera2019gramatica}, \cite{sasakidivide}).

\begin{table}[h!]  
  \begin{center}
     \caption{Examples of syntactic constructions in Nawatl language. More frequent structures (in relationship with the verbs) are indicated in bold.}\label{tab:EXEMP_GRAM}
\resizebox{0.49\textwidth}{!}{%
    \begin{tabular}{|l|c|c|}
     \hline
    \bf Structure & \textbf{Example} & \textbf{Translation} \\
    \hline
   \bf VSO & Kitta tlakatl kalli & Sees (a) man (a) house \\ \hline
   \bf VO & Kitta kalli & (He/she) sees (a) house \\ \hline
   \bf VS & Kitta tlakatl & Sees (him/her/it) (a) man \\ \hline
   \bf VOS & Kitta kalli tlakatl
 & Sees (a) house (a) man \\ 
   \hline \hline
   SV & Tlakatl kitta & (A) man sees (him/her/it) \\ \hline
   SVO & Tlakatl kitta kalli & (A) man sees (a) house \\ \hline
   SOV & Tlakatl kalli kitta & (A) man (a) house sees \\
  \hline
  \end{tabular}
  }
  \end{center}
\end{table}

\section{Context-Free Grammars}
\label{sec:GNC}
A Context-Free Grammar (CFG) is a type of formal grammar used to describe the syntax of formal languages, especially in the syntactic analysis of programming languages and natural languages \cite{hopcroft2006automata}.

Formally, a CFG is a quadruple:
\[
G = (V, \Sigma, R, S)
\]
where:

\begin{itemize}
    \item $V$ is a finite set of \textbf{non-terminal symbols}.
    \item $\Sigma$ is a finite set of \textbf{terminal symbols}, such that $V \cap \Sigma = \emptyset$.
    \item $R$ is a finite set of \textbf{production rules}, of the form:
    \[
    A \rightarrow \alpha
    \]
    where 
    $A \in V$, $\alpha \in (V \cup \Sigma)$, 
    and 
    $(V \cup \Sigma)$
    represent all possible strings
    (of length 0 or more) formed with symbols from $V$ and $\Sigma$.

\item $S \in V$ is the start symbol.
\end{itemize}

A grammar is called context-free because the rules are applied without considering the context in which the non-terminal symbol appears.

\section{A Context-Free Micro-grammar for Nawatl}
\label{sec:GNC_proposed}

In this section, we introduce a new context-free micro-grammar for Nawatl.
This grammar is considered as a micro version primarily because it avoids the use of recursive production rules.
Similarly, only the first three grammatical persons (I, you, he/she, it) are included, verbs being limited to the singular form and present tense.

\subsection{A first context-free micro-grammar for Nawatl: $\mu$\textsc{gnaw}$\oplus$0}

Our initial approach to Nawatl grammar (see Table \ref{tab:gram0}) $\mu$\textsc{gnaw}$\oplus$0, is inspired by the grammatical frameworks developed for Indo-European languages.
Given that Nawatl belongs to the families of indigenous American languages —which are linguistically distant from the Indo-European family— such a model results in a limited and reductive representation of the language.
Nevertheless, the objective here is not to provide an accurate model of real Nawatl, but rather to illustrate that the conventional structure: 

$P\rightarrow$ noun phrase ($N$)  verb phrase ($V$), 

\noindent where $V$ may also contain $N$ is insufficient to capture the deeper syntactic organization of Nawatl, which is centered around the verb.
We therefore focus on a classical model consisting of two primary phrase types: the noun phrase ($N$) and the verb phrase ($V$).
These phrases may include additional elements (particles, nouns, and prefixes) that are translated into our grammatical categories as temporal and quantifying adverbs ($ADV_T, ADV_Q$), adjective nouns ($ADJ$), personal ($PP$) and person markers ($PV$), possessive markers ($POS$), and negation ($NEG$).
Although derived from the grammars of families of $\tau$-languages, this preliminary version nonetheless enables the generation of some basic Nawatl structures.

For the sake of simplifying our approach, in the two micro-grammars we will use only the singular grammatical persons. The plural will be left for a later work.

\label{sec:grammar}

\begin{table}[h!]
\lstset{
   basicstyle=\small,
  mathescape=true
}

\begin{lstlisting}
P -> ADV$_{T}$ (N|V)
N -> ADJ (ART_|POS)+n 
V -> N NEG PV$_{3}$+v ADV$_Q$
V -> PP$_{i}$ NEG PV$_{j}$+v ADV$_{Q}$; i,j=1,2,3; i=j

ADV$_{Q}$ -> miyak|tlawel|vide     # a lot|too much|vide
ADV$_{T}$ -> naman|axcan|axan|vide # now/this day/today|vide
ADJ -> tomawak|kualtzin|vide  # fat|nice|vide
ART -> se|ni|vide             # one|the,this|vide 
POS -> no|mo|i             # my|your|his,her,its
PP$_i$ -> na|ta|ya            # I,me|you|he,she,it
PV$_j$ -> ni|ti|vide             # I|you|he,she,it|vide
NEG -> amo|axkeman|vide       # no|never|vide

n -> siwatl|miston|elotl|xokotl|tochin|yolkatl|nakatl
   # woman|cat|corn|fruit|rabbit|animal|meat|...
v -> nehnemi|kwa|kaki 
   # to walk|to eat|to listen|...

$\oplus$ = concatenation $\emptyset$ = null $\_$ = space
\end{lstlisting}

\caption{Grammar $\mu$\textsc{gnaw}$\oplus$0.\label{tab:gram0}}
\end{table}

We present here a few examples of the production rules of the micro-grammar, in generative form.
For instance, the noun phrase $N$ can generate:

{
\begin{itemize}
\item $ADJ ~ ART ~ n$:  \text{Yehyektsin ni/ne siwatl} \\
\textit{\# Beautiful [is] this/that woman}
\item $ADJ ~ POS ~ n$: \text{Tomawak motoch} \\
\textit{ \# Fat [is] your rabbit}
\end{itemize}
}
Similarly, the verb phrase $V$ could yield productions such as the following:
{
\begin{itemize}
\item $PP_{1,2,3} ~~ PV_{1,2,3} \oplus v$: na/ta/ya ~ ni/ti/\_/nehnemi \\ 
\textit{\# I am/you are/he/she is (who) I/you/he/she walks}
\item $POS_1 \oplus n ~ PV_3 \oplus v$: nomiston \_nehnemi \\
\textit{\# my\_cat ~ it\_walks}
\item $POS_2 \oplus n ~ NEG ~ PV_3 \oplus v$: momiston amo \_nehnemi \\
\textit{\# your\_cat it doesn't walk}
\item $PP_1 ~ NEG ~ PV_1\oplus v ~ ADV_Q ~ n$: 
\text{na amo nikkwa miyak xokotl} \\
\textit{\# It's me (who) I don't eat a lot [of] fruit}
\item $ADV_T ~ ADJ (ART~|POS)\oplus n ~ NEG ~ PV_3 \oplus v ~ ADV_Q ~ n$:
\text{axan tomawak ni /no/toch(in) amo \_(tla)kwa tlawel elotl} \textit{\# Now fat the/my rabbit doesn't  eat too much corn} 

\end{itemize}
}

\subsection{Computing the Number of Productions in $\mu$\textsc{gnaw}$\oplus0$}

\section{$\pi$-\textsc{yall}-\textsc{ia}, an augmented Artificial Nawatl corpus}
\label{sec:artificial_corpus}

We observed that the grammar $\mu$\textsc{gnaw}$\oplus$0 is capable of generating a large number $\mathfrak{F}^{0}$ of grammatically acceptable productions.  
However, although the value $\mathfrak{F}^{0}$ is large, the sentences produced represent only a tiny fraction of the number that could be produced using recursive grammars.  
Such recursive grammars, however, fall outside the scope of this article.

Nevertheless, the number of sentences $\mathfrak{F}^{0}$ remains sufficient to enable the artificial enrichment of the $\pi$-\textsc{yalli} corpus. 
To achieve this, we must establish certain restrictions in order to produce not only grammatically correct but also semantically acceptable sentences. 
Indeed, we prefer not to accept sentences such as:

``\textit{The big corn-cob eats a lot of rabbit}'' or ``\textit{The fruit dreams too much}''

\noindent because, although they belong to $\mu$\textsc{gnaw}$\oplus$0, they are not semantically realistic. We must therefore implement filters.

\subsection{Semantic Filters}

The artificial corpus derived from the grammar presented should contain sentences that can be effectively used for training static LLMs. 
In order to retain only sentences that are both grammatically correct and semantically acceptable, we propose to introduce:

\begin{itemize}
    \item A filter that attempts to approximate semantics using supervised classification methods.
    \item A filter based on the association between verbs and animate/inanimate nouns.
    \item No repeated nouns in the same sentence.

\end{itemize}

The semantic filter based on supervised learning is computationally greedy and will be implemented in future work.  

The animate/inanimate filter enables the elimination, at low computational cost, of a significant subset of sentences that lack reasonable semantics. Some examples of the filter can be seen in appendix $1$. 

This approach enabled us to augment the $\pi$-\textsc{yalli} corpus with an additional number of N tokens, while still respecting a plausible Nawatl grammar.

From the grammar $\mu$\textsc{gnaw}$\oplus$0, we generated sets of $\mathfrak{F}^{0}$ sentences.  
These sets, once appropriately filtered, yielded one version of the $\pi$-yall-\textsc{ia}$\oplus$0* corpus, containing respectively $\mathfrak{F}^{0}*$ sentences.  
The semantic approximation filter associates nouns with an animate referent (e.g. animals, people) to verbs expressing appropriate actions (e.g. to walk, to eat, to fly), in contrast to inanimate referents (e.g. things, artifacts, phenomena) and their corresponding verbs.

This procedure allowed us to retain only a set of $\mathfrak{F}^{0}*$ grammatically correct and semantically acceptable sentences, resulting in an artificial and usable corpus of Nawatl sentences.
In appendix $2$ we present several examples for our grammar.

Accordingly, $\mu$\textsc{gnaw}$\oplus$0, we obtained:
    $\mathfrak{F}^{0}* \approx$ 807K sentences with both filters applied and using the terms of our knowledge base.

\begin{table}[h]
\center
\begin{tabular}{|c|c|c|c|c|c|c|c|c|c|}
 \hline
  $N$ & $V$ & $ADV_Q$ & $POS$ & $ART$ & $ADV_T$ & $ADJ$ & $PP$ & $NEG$ &$\mathfrak{F}^0*$  \\ \hline
  26 & 16 & 5 & 3 & 3 & 7 & 3 & 3 & 3 & $807,093$  \\ 
  \hline 
\end{tabular}
\caption{Number of elements for $\mu$\textsc{gnaw}$\oplus$0. Appendix 3 shows our knowledge base}
\label{T7}
\end{table}

\subsection{Merging artificial sentences and the Nawatl $\pi$-{\sc yalli} corpus}
\label{sec:corpus}

The idea that emerges now is the combination of an ``authentic'' corpus such as $\pi$-\textsc{yalli} with an artificially generated corpus. We think that this union may be beneficial for teaching Language Models (whether static or LLM). In particular, we have focused on static LM training. 
After combining the two corpora, the new corpus goes through an elementary process of unification of spellings \cite{MICAI-piyalli-unigraph}, in order to standardize the use of characters of the different variants of Nawatl\footnote{For example, changing c for k, hu for w, the elimination of double consonants, the use of lowercase letters, avoiding accents, etc.}.

We now present some characteristics of the --authentic-- $\pi$-{\sc yalli} Nawatl corpus \cite{piyalliTALN}, that we have used for our semantic similarity experiments\footnote{The $\pi$-yalli corpus may be founded on the website: \url{https://demo-lia.univ-avignon.fr/pi-yalli}}
The corpus is heterogeneous in terms of categories and linguistic variants in Nawatl (see Table~\ref{tab:statcorpus}), and contains a relatively small number of tokens ($\approx$~6.1M) and sentences ($\approx$ 333K)\footnote{$\pi$-yalli in its version v1.8. A new, revised version v1.9 with more documents has been also tested.}.  
This makes it useful for training static LLMs, but clearly inadequate for training dynamic LLMs. 
Indeed, it has been reported that dynamic LLMs require in the order of 10M-100M tokens to achieve stable embeddings~\cite{micheli2020importancepretrainingdatavolume}.

For this reason, we decided to unify the $\pi$-\textsc{yalli} corpus with the filtered artificial sentences generated by the grammars.
The new corpus, $\pi$-\textsc{yall}-\textsc{ia}$\oplus0$, was employed to train FastText static embeddings with the objective of improving performance on the task of semantic similarity between sentences.

\begin{table}[h!]
\centering
\footnotesize
\resizebox{1\textwidth}{!}{%
 \begin{tabular}{|c|r|r|r|r|}
  \hline
  \bf Category & \bf \# docs & \bf Variants of Nawatl& \bf Tokens & \bf \% Corpus \\ \hline
  AGR & 3 & cen(2),hua(1) & 7 828 & 0.13 \\ \hline
  COS & 3 & cen(1),hua(1),cla(1) & 40 983& 0.67 \\ \hline
  ECO & 1 & cen(1) & 16 777&0.27 \\ \hline
  EDU & 67& cen(66),hua(1) & 276 763&4.52 \\ \hline
  HIS & 54 & cla(47),cen(6),pue(3)& 645 406& 10.54 \\ \hline
  LEG & 20 & cen(8),cla(3),pue(1),hid(2),hua(4) & 341 945& 5.58 \\ \hline
  LIN & 9 & cen(6),hua(2),cla(1) & 368 511& 6.02 \\ \hline
  LIT & 52 & cen(31),pue(10),gue(4)& 881 369&14.39 \\ \hline
  MED & 4 & cen(2),hua(2) & 14 248&0.23 \\ \hline
  MUS & 5 & cen(5) & 4 306 &0.07 \\ \hline
  PHR & 49 & cen(42), mix(7) & 9 259 &0.15 \\ \hline
  POE & 11 & cen(9), mix(2) & 5 647 &0.09 \\ \hline
  POL & 2 & mor(1),cen(1)& 1 082 &0.02 \\ \hline
  REL & 29 & cla(14),cen(4),pue(5),gue(1),oax(3),hua(2)& 3 308 363& 54,05 \\ \hline
  TEC & 1 & cen(1) & 518 & 0,01 \\ \hline
  WIK & 4 298& mixture of variants &194 292& 3,17 \\ \hline
  \bf TOTAL&\bf 4 608 &cen(185),cla(66),pue(19),hua(13),hid(2),oax(3),mor(1) &$\approx$\bf 6 121 000 &\bf 100.0 \\ \hline
  \multicolumn{2}{|l|}{\bf 333K Sentences} & \multicolumn{3}{|l|}{ }\\ 
  \hline
\end{tabular}
}
\caption{\small 
 Statistics on the documents in the $\pi$-\textsc{yalli} v1.8 corpus. cen = central, cla = classical, pue = Puebla, gue = Guerrero, oax = Oaxaca, mor = Morelos, hua = Huasteca, hid = Hidalgo; in parentheses, the number of documents per variety. In several cases, there may be a mixture (mix). 
 AGR: Agriculture; COS: Cosmovision; ECO: Economy; EDU:Education; HIS:History; LEG: Legal documents; LIN:Linguistics; LIT:Literature; MED:Medicine; MUS:Music; PHR:General sentences; POE:Poetry; REL=Religion; TEC:Science and Technology; WIK:Wikipedia.
 \label{tab:statcorpus}}
\end{table}

\section{A sentence-level semantic similarity task}
\label{sec:experiments}

In this section, we present a protocol to evaluate our approach. Semantic similarity is a standard task in Natural Language Processing (NLP), involving the evaluation of various models (statistical models, neural networks, etc.) through standardized evaluation protocols \cite{francis-landau-etal-2016-capturing}.  
In particular, we focus our attention on the task of semantic similarity between a reference sentence and a set of candidate sentences.  
This gives rise to a ranking of candidates that can be compared against a human-generated ranking.  

This is the evaluation protocol used in \cite{piyalliTALN}, which we adopt here by training embeddings to assess the impact of learning based on both existing and artificially generated corpus.

\color{black}

\begin{figure}[h!]
  \centering
  \includegraphics[width=0.7\linewidth]{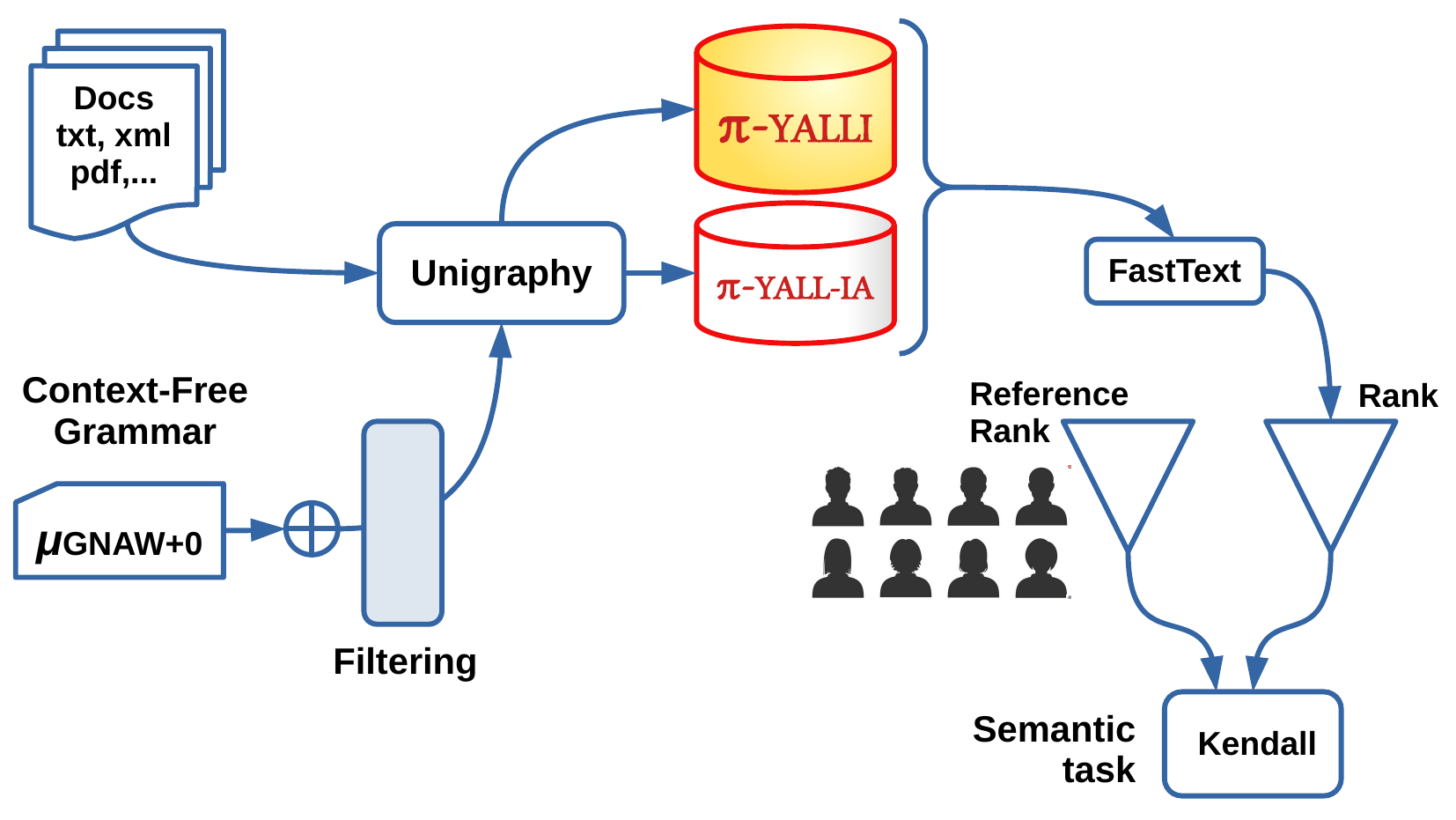}
    \caption{Outline of the Sentences Semantic Similarity task.}
    \label{fig:flow}
\end{figure}

Figure~\ref{fig:flow} illustrates the complete experimental pipeline we designed, from the construction of the corpus (including authentic documents or generated from the formal grammar) to the evaluation of the semantic task via a reference ranking produced by human annotators, compared to the ranking generated by FastText.

We did not expect to achieve high performance on the semantic task using only the artificial corpus generated by the $\mu$\textsc{gnaw}$\oplus0$ grammar.  
Rather, our goal was to enrich the $\pi$-\textsc{yalli} corpus with these new artificial datasets.  
Our hypothesis is that the $\mu$\textsc{gnaw}$\oplus$0 corpus should contribute to improving the performance of static language models.

The extended corpora $\pi$-yall-\textsc{ia}$\oplus0$ were thus used to train FastText language models \cite{bojanowski-etal-2017-enriching} from scratch.  
The resulting embeddings were then evaluated on a semantic sentence similarity task.  
The semantic similarity task used is the following: given a set of reference sentences $r_i$, $i=1,2,...,R$ (without context), where the reference of each sentence is associated with a list L of 5 candidate sentences, the goal is to sort the list L according to the semantic similarity to its corresponding reference sentence.

A total of $R=30$ references were written in Spanish by an annotator. These $R$ sentences were divided into 6 blocks of 5 sentences each. Each block was assigned to an annotator who drafted $C=5$ candidate sentences with varying degrees of semantic similarity relative to the original sentence. 
All sentences (references and candidates) were translated into the Central Nawatl variant by a bilingual Nawatl speaker (Spanish-Nawatl).

The basic statistics for this task are as follows in Table \ref{tab:stats_task}.
\begin{table}[h!]
    \centering
\begin{center}
\resizebox{0.6\textwidth}{!}{%
 \begin{tabular}{|c|r|r|r|c|}
 \hline
  & \bf Sentences & \bf Tokens & \bf Types &\bf Tokens per Sentence  \\ \hline
  \bf References & 30 &246 & 189 & 8.20 \\ \hline
  \bf Candidates & 150 &1026 & 599  & 6.84 \\ \hline
 \end{tabular}
 }
\end{center}
    \caption{Statistics of the semantic task.}
    \label{tab:stats_task}
\end{table}

Appendix shows two examples of reference–candidate sentence blocks that are included in this reference ranking ($R_R$).

The reference ranking $R_R$ is produced by humans, and the goal of the task is to measure how closely the rankings $R_M$ generated by various language models approximate the reference ranking $R_R$.
The similarity measure between rankings is the Kendall's $\tau(R_R, R_M)$.

\subsection{The LLMs employed}

In this paper we tested the following five LLMs models via APIs\footnote{
The API accesses to LLMs were performed on 09/06/2025 from 16h-20h GMT. The interactive ones (Copilot, Grok and Claude) were performed on June 9-10/2025.}: ChatGPT-4 mini API; Gemini-2.5-flash-preview-05-20 API\footnote{\url{https://deepmind.google/models/gemini/flash/}}; DeepSeek-V3-0324 API\footnote{\url{https://api-docs.deepseek.com/news/news250325}}; Llama-3.1-70B-Instruct API\footnote{\url{https://huggingface.co/meta-llama/Llama-3.1-70B-Instruct}} and Mistral-large-latest API\footnote{\url{https://mistral.ai/news/mistral-large}}. In an interactive user mode we employed three models: Copilot (\url{https://www.microsoft365.com}), Grok 3 (\url{https://grok.com}) and Claude 3.7 (\url{https://claude.ai}).

The prompt used was the following:\footnote{
The prompt was written in French: <<{ \it
\noindent \`A partir de la phrase nahuatl: ``[référence.]'' triez par ordre sémantique, de la plus proche à la plus lointaine, les cinq phrases suivantes: ``[candidate$_1$.]''; ``[candidate$_2$.]''; ``[candi\-date$_3$.]''; ``[candidate$_4$.]''; ``[candi\-date$_5$.]''.  Ne donnez pas des rankings de phrases autres que celles proposées.~}>>}
{ \it
\noindent Given the Nawatl sentence ``[reference.]'' rank the following five sentences semantically, from the closest to the furthest in meaning from the original sentence:
``[candidate$_1$.]''; ``[candidate$_2$.]''; ``[candidate$_3$.]''; ``[candidate$_4$.]''; ``[candidate$_5$.]''.
Do not give ranking of other sentences than those provided.} The references and candidates were written in Central Nawatl.

\subsection{Results and Discussion}

Table~\ref{tab:tau_phrases_LLM} summarizes all our results.
We conducted tests with version 1.8 of the $\pi$-{\sc yalli} corpus (6.12M tokens) and with a new $\pi$-{\sc yalli} version 1.10 (with $\approx$ 6.63M tokens).
This will, of course, allow us to verify the hypothesis that the greater the number of tokens, the better the models learn.

As shown in this Table, the best result was obtained using the filtered $\pi$-\textsc{yall}-\textsc{ia}$_{1.10}\oplus$0 corpus, which yielded a Kendall's $\tau$ = {\bf 0.527}. This places the algorithm FastText model trained with an expanded corpus in 3rd place in the ranking, only behind the LLMs Gemini 2.5 and Claude 3.7.

\begin{table}[ht!]
  \centering
  \resizebox{0.49\textwidth}{!}{%
  \begin{tabular}{|c|c|}
  \multicolumn{2}{c}{\bf Sentences Semantic Similarity Task}  \\ 
\hline
  {\bf Algorithm} & Kendall's $\tau$\\ \hline
  Gemini 2.5-flash-preview-05-20 API &\bf 0.693  \\ \hline 
  Claude 3.7 & 0.607 \\ \hline


\color{blue} \textit{\textbf{FastText / $\pi$-{\sc yalli} 1.10 $\cup$ $\mu${\sc gnaw}$\oplus$0}}  &\color{blue} \textit{\textbf{0.527} } \\ \hline
  
\color{blue} \textit{FastText / $\pi$-{\sc yalli}} 1.10 &\color{blue} \textit{0.513} \\  \hline

  DeepSeek V3-0324 API & 0.513 \\ \hline 
  Grok 3& 0.487  \\ \hline

  \color{blue} \textit{FastText / $\pi$-\textsc{yalli 1.8} $\cup$ $\mu${\sc gnaw}$\oplus$0}&\color{blue} \textit{0.487} \\ \hline

  Copilot & 0.473  \\ \hline

\color{blue} \textit{FastText / $\pi$-\textsc{yalli 1.8}}&\color{blue} \textit{0.467} \\ \hline

  Mistral-large-latest API& 0.387 \\ \hline
  ChatGPT-4 mini API & 0.367 \\ \hline
  Llama-3.1-70B-Instruct API& 0.233  \\
  \hline
  \end{tabular} 
  }
\caption{Sentence Semantic Similarity, Kendall’s $\tau$: LLMs vs FastText (SkipGram model on 300 dimensions) without the four stopwords: \textit{iwan, in, tlen}, and \textit{ipan}. The maximum values obtained in {\it Leave-one-out} tests are shown. Our models are in blue.}
 \label{tab:tau_phrases_LLM}
\end{table}


On the other hand, if we compute the ratios $\rho_{0}$ representing the proportion of artificial text (tokens and sentences) produced by our grammar with respect to the texts available in the $\pi$-\textsc{yalli} v1.10 corpus, we obtain:

$\rho_0$ (tokens) = $\frac{4.6M}{6.63M} \approx$ 70\%;

$\rho_0$ (sentences) = $\frac{809K}{417K} \approx$ 193\%.


\section{Conclusions and Future works}
\label{sec:conclusions_future}

The generative use of the proposed $\mu$\textsc{gnaw}$\oplus$0 grammar enables the creation of a large number $\mathfrak{F}^{0}$ of sentences, that may be considered as a --very large--  artificial corpus.  
A semantic filtering process was applied to this artificial corpus in order to reduce the universe to $\mathfrak{F}^{0}*$ sentences.
Subsequently, this artificial corpus was concatenated with the authentic $\pi$-\textsc{yalli} corpus.  
The two corpora together contain a total of approximately 11 million Nawatl tokens.

The approach to Nawatl grammar using a CFG micro-grammar in generative mode, seems to help the learning process of a FastText-like static model. The resulting embeddings better capture the structure of the Nawatl language. However, there are still difficulties and challenges in relation to the deep structure of an agglutinative language, morphophonological modifications of roots by composition, and features specific to dialect variation. 
Even though different syntactic structures can be produced and thus new grammatical sentences generated, the complexity of syntactic relationships must be explored in depth. 

These difficulties are lessened in the case of processing isolating or moderately synthetic languages such as Indo-European languages.
Nawatl, as a polysynthetic language, is a complex morphological case that requires, even for humans\cite{kelly2014acquisition}, the ability to identify roots on which to incorporate multiple affixes, but also the ability to comprehend the rules of composition and, above all, phonetic and morphological modifications. To all this, we must add the dialectal diversity of the forms.
Thus, in our case the problem is to achieve a morphological regularity and not a morphological complexity as the relevant factor for learning, since this kind of language may be relatively regular in the template sequence used \cite{kelly2014acquisition}.

In future work, we wish to construct a CFG grammar that is closer to Nahuatl grammar, not only based
on Indo-European grammars.
We will seek to increase the number of $n$ and $v$ elements, include additional grammatical persons, verb tenses, plurals, rhetorical connectors, and improve the semantic filter (to avoid effects of combinatorial explosion).
This contribution adds to research that, in spite of the fact that these are highly complex and low-resource languages, recognizes the broader possibilities for computational development of languages of the Uto-Aztecan family \cite{mager2024low}, \cite{eskander2019unsupervised}. 

We also aim to introduce an extended recursive grammar to try of modeling the more frequent structures produced by native Nawatl speakers more realistically and, additionally, to evaluate the contribution of the artificially generated corpus to Detection of Sentiment Analysis, Automatic Text Summarization \cite{torres14} and Named Entities recognition task.



\section*{Acknowledgements}
This research is funded by the Intermedius School, Université d'Avignon grant for a doctoral PhD thesis (France), and partially funded by the Université d'Avignon, Laboratoire Informatique d'Avignon (LIA), and the Agorantic research program (France). M.F.-S. and L.Q.T are also grateful for the kind hospitality of the LIA and Intermedius, where they participated in implementing this research project.

\bibliographystyle{apalike}
{\small
\bibliography{references}}


\section*{Appendix 1: Filtering of animate/inanimate elements}
\label{sec:appendix1}

Our proposed filtering works by tagging nouns and verbs with one of the tags already described. 
To illustrate how this filter works, we present some examples of suitable verbs and nouns.

{The concrete implementation of the filter is as follows: a sentence is constructed using only nouns associated with verbs of the same type ($n$ animate/$v$ animate or $n$ inanimate/$v$ inanimate), avoiding the mixing of animate/inanimate ones.}

 Nouns:
\begin{itemize}
    \item nantzin = mother (\textbf{animate})
    \item mihkailwitl = day of the dead (\textbf{inanimate})
    \item tatzin = father (\textbf{animate})
    \item mapachin = raccoon (\textbf{animate})
    \item kuawtli = eagle (\textbf{animate})
\end{itemize}

 Verbs:
\begin{itemize}
    \item miki = to die (\textbf{animate})
    \item ixpoliwi = to disappear (\textbf{animate/inanimate})
    \item pia = to have (\textbf{animate/inanimate})
    \item itta = to see (\textbf{animate})
    \item chihua = to make (\textbf{animate})
\end{itemize}

The rule says that if the labels match, the sentence can be generated, but if the label is animate/inanimate, then a sentence with a live and nolive can be generated. That is, father (\textbf{tatzin}) and mother (\textbf{nantzin}) can die (\textbf{miki}), but not on the "Day of the Dead" (\textbf{mihkailwitl}), however, everyone can disappear (\textbf{ixpoliwi}).

\section*{Appendix 2: Examples of artificial sentences generated by our micro-grammar}
\label{sec:appendix2}

This appendix shows some examples of artificial sentences generated by the grammar $\mu$\textsc{gnaw}$\oplus$0. 
This grammar generates 807k sentences after passing through all the necessary filters. Here are some examples with their approximate translations:
\begin{itemize}
    \item \textit{Aman weyi ni ichpochtli amo toka miakeh.} |
    Now lady dooesn't bury several ones.
    \item \textit{Aman weyi ni ichpochtli amo ahsikamati nochi.} |  Now big lady doesn't understand everything.
    \item \textit{Axkan weyi mokoltzin amo maka miyak.} |
    Now grandfather doesn't give much.
    \item \textit{Axkan weyi mokoltzin axkeman kaki achi.} | 
    Now grandfather never listens enough.
    \item \textit{Tonalli weyi temachtiani ixpantilia nochi.} | Day great teacher shows [to somebody] everything. 
\end{itemize}

\newpage

\section*{Appendix 3: Prolog Knowledge Base of $\mu$\textsc{gnaw}$\oplus0$}
\label{sec:appendix3}

\begin{table}[h!]
\centering
\small
\resizebox{1\textwidth}{!}{%
 \begin{tabular}{|c|c|c|c|c|c|}
  \hline
  \multirow{6}{*}{\bf NOUNS}   & \bf ichpochtli & \bf tlakanechikolli & \bf mapachin & \bf yolkatl & \bf telpochtli \\
                              & \it maiden & \it group of people & \it raccoon & \it animal & \it young man \\ \cline{2-6}
                              & \bf kuawtli & \bf koltzin & \bf ilamah & \bf weweh & \bf cihuamichin \\
                              & \it eagle & \it grandfather & \it old woman & \it old man & \it mermaid \\ \cline{2-6}
                              & \bf tlakatl & \bf tlamatini & \bf tonatih & \bf sentilistli & \bf momachtiani \\ 
                              & \it man & \it wise man/woman & \it sun & \it family & \it student \\ \cline{2-6}
                              & \bf temachtiani & \bf posolli & \bf xochitl & \bf siwatl & \bf tlahtolli \\
                              & \it teacher & \it pozole & \it flower & \it woman & \it speech \\ \cline{2-6}
                              & \bf tatzin & \bf nantzin & \bf tiotzin & \bf tototl & \bf coyotl \\ 
                              & \it father & \it mother & \it god & \it bird & \it coyote \\ \cline{2-6}
                              & \bf tateh & & & & \\ 
                              & \it sir & & & & \\ \hline \hline
  \multirow{3}{*}{\bf VERBS} & \bf toka & \bf kaki & \bf ahsikamati & \bf itta & \bf chihua \\ 
                              & \it to bury & \it to listen & \it to understand & \it to see & \it to make \\ \cline{2-6}
                              & \bf chiya & \bf pia & \bf mati & \bf maka & \bf ixpantilia \\ 
                              & \it to look & \it to have & \it to Know & \it dar & \it to show [sth] to [sb] \\ \cline{2-6}
                              & \bf machtia & \bf welitta & \bf neki & \bf tlasohtla & \bf paka \\ 
                              & \it to learn [sth] to [sb] & \it to like & \it to want & \it to love & \it to wash \\ \cline{2-6}
                              & \bf paktia &  &  &  &  \\ 
                              & \it to make happy &  &  &  &  \\ \hline \hline
\multirow{3}{*}{\bf QUANTITATIVE AND LOCATIVE MARKERS} & \bf ahko & \bf miyak & \bf tlakpak & \bf ompa & \bf aman \\
                              & \it up & \it a lot of & \it high & \it there & \it now \\ \cline{2-6}
                          & \bf axkan & \bf miakeh & \bf nochi & \bf seki & \bf achi \\
                          & \it now & \it many/much & \it all & \it some & \it enough \\ \cline{2-6}
                          & \bf nepa & \bf ompa & \bf oncan & \bf axan & \bf nama \\ 
                          & \it that & \it there & \it over there & \it today & \it today \\ \cline{2-6}
                          & \bf niman & \bf tonalli & \bf yalwa &  &  \\
                          & \it after/then & \it day & \it yesterday &  &  \\ \hline \hline
\multirow{2}{*}{\bf ADJECTIVE NOUNS }  & \bf weyi & \bf istak & \bf miyak &  &  \\
                                 & \it big & \it white & \it many/much &  &  \\ \hline \hline
\multirow{2}{*}{\bf POSSESSIVE MARKERS }  & \bf no\_ & \bf mo\_ & \bf i\_ &  &  \\
                                 & \it my & \it your & \it his/her/its &  &  \\ \hline \hline
\multirow{2}{*}{\bf SUBJECT MARKER OF VERB }  & \bf ni\_ & \bf ti\_ & \bf \_ &  &  \\
                                 & \it I & \it you & \it he/she/it &  &  \\ \hline

\end{tabular}
}
\end{table}

\end{document}